\documentclass[pmlr]{jmlr}

\usepackage{microtype}
\usepackage{booktabs}
\usepackage{comment}
\usepackage{tcolorbox}
\usepackage{tikz}
\usepackage{fvextra}
\usepackage[utf8]{inputenc}
\usepackage{array}
\fvset{breaklines=true, breaksymbolleft={}, fontsize=\footnotesize}
\usetikzlibrary{arrows.meta,positioning}

\tcbuselibrary{skins,breakable}

\definecolor{odred}{RGB}{178,24,24}
\definecolor{archblue}{RGB}{29,78,216}

\newif\ifshowopen
\showopenfalse

\ifshowopen
  \newtcolorbox{opendecision}[1]{
    colback=odred!5, colframe=odred, colbacktitle=odred, coltitle=white,
    fonttitle=\bfseries\small, fontupper=\small, title={Open decision:~#1},
    breakable, boxrule=0.6pt, left=6pt, right=6pt, top=3pt, bottom=3pt
  }
  
\else
  \excludecomment{opendecision}
  
\fi

\jmlrproceedings{}{FrontierOR Challenge White Paper}
\jmlrvolume{}
\jmlrpages{}
\jmlryear{2026}
\jmlrworkshop{Version 1.0, 20 November 2026}

\title[FrontierOR Challenge]{The FrontierOR Challenge: Agentic Algorithm Design for Operations Research}

\author[Wang \emph{et al.}]{%
  \mdseries Organizers\\
  \Name{Hai Wang}, \Name{Bretislav Hajek}, \Name{Chonghe Jiang}, \Name{Minwei Kong},\\
  \Name{Ao Qu}, \Name{Huatian Gong}, \Name{Bo Dong}\\[6pt]
  \Name{Cynthia Barnhart}, \Name{Jinhua Zhao}\\[6pt]
  \Email{frontieror.competition@gmail.com}}

\jmlrauthors{H. Wang \emph{et al}}

\begin{document}

\maketitle

\begin{abstract}
Operations research (OR) relies on algorithms that experts tailor to each problem: a new routing, scheduling or network design problem can take weeks of specialist work before it is solved well at scale. Large language model (LLM) agents now write, test and improve code, but whether they can design algorithms that scale to realistic instances is open: in the FrontierOR benchmark's one-shot evaluation, the strongest model matches Gurobi, with an objective within 1\% of Gurobi's and no slower, on only 39\% of the large instances. The FrontierOR Challenge asks teams to build agentic workflows that design, test, and improve algorithms for many optimization problems at once. Stage~1 releases 60 of the benchmark's 180 audited problems, spanning six application categories and five formulation types, in weekly rounds of 15; Stage~2 introduces new problems outside the benchmark, with unseen instances. Two tracks are ranked separately: in the \emph{Main track} we run the submitted algorithms in a network-isolated sandbox with open-source dependencies only; in the \emph{Self-hosted track} teams solve the instances on their own machines, with any software, within a timed window. Every solution is checked for feasibility, and a feasible solution scores the ratio of its objective to a frozen known reference, averaged over hidden instances.
\end{abstract}

%=====================================================================
\section{Introduction}
%=====================================================================

The \emph{FrontierOR Challenge}\footnote{\url{https://frontieror-challenge.com}} asks teams to build agentic workflows that design, test, and improve algorithms for optimization problems, instead of hand-tuning a single algorithm. The main scientific question is the following: \emph{``Can an LLM agent design algorithms that are feasible, good and fast on instances it has never seen, across many optimization problems and under a fixed compute budget?''}

Optimization problems span many categories, such as routing, scheduling and network design. For each of them, the OR literature studies specialized algorithms: decompositions, branch-and-price, problem-specific heuristics. General-purpose solvers help, but a correct formulation often still requires a deep understanding of the domain. Within a one-hour time limit, Gurobi does not reach optimality on 46\% of the large FrontierOR instances \citep{kong2026frontieror}. LLM-driven program search has started to change this. FunSearch \citep{romera2024funsearch} and AlphaEvolve \citep{novikov2025alphaevolve} found new constructions by evolving programs, and EoH \citep{liu2024eoh} and ReEvo \citep{ye2024reevo} evolve heuristics for combinatorial problems. In optimization modeling, AlphaOPT builds and refines a library of reusable insights based on solver feedback to guide LLMs in formulating new problems~\citep{kong2025alphaopt}. Benchmarks measure the gap that remains: CO-Bench \citep{sun2026cobench}, HeuriGym \citep{chen2026heurigym}, ALE-Bench \citep{imajuku2025alebench} and FrontierOR \citep{kong2026frontieror} all find that agents trail expert algorithms and commercial solvers, and that consistency across problems is the weakest point. In FrontierOR's one-shot setting, a model writes one program per problem from its statement, with a few self-debugging rounds on execution errors. In the latest evaluation, the strongest model, Claude Fable 5, returns a feasible solution within 1$\%$ of or better than Gurobi's reference objective, in no more than Gurobi's wall-clock time, on 39$\%$ of the large instances across all 180 problems. Results in this competition may differ, since the Main track allows only open-source libraries, whereas in FrontierOR the generated algorithms could call Gurobi.

The competition captures a practical scenario: a practitioner faces a new decision problem with a written statement, a data format, and a deadline, and needs an algorithm that works on unseen instances. Each team receives the problem definitions, builds a pipeline that turns each one into an algorithm, and is scored on hidden instances. The competition runs in two stages: a qualification on 60 problems from the FrontierOR benchmark, released 15 per week, and a final on new problems, and in two tracks: the Main track, where we run the submitted algorithms under fixed compute with open-source dependencies, and the Self-hosted track, where teams solve the instances on their own infrastructure with any software, including commercial solvers.

We expect the competition to contribute (1)~a public comparison of agentic algorithm design methods under one protocol, with independent checkers and hidden instances; (2)~evidence for practitioners on whether an agent can replace weeks of expert work on a new problem; (3)~new problems and instances for the OR community, through a call for contributions; and (4)~released problems, checkers and descriptions of the top teams' methods after the competition.

%=====================================================================
\section{Related Competitions}
%=====================================================================

Earlier competitions at the intersection of machine learning and OR focused on one part of the pipeline. NL4Opt \citep{ramamonjison2022nl4opt} asked for formulations of small linear programs from text. ML4CO \citep{gasse2022ml4co} replaced heuristic components inside a MIP solver on three datasets. The DIMACS Implementation Challenges\footnote{\url{http://dimacs.rutgers.edu/programs/challenge/}} and the ROADEF/EURO Challenge\footnote{\url{https://www.roadef.org/challenge/}} pose one problem per edition to human algorithm designers, as do the AtCoder Heuristic Contests,\footnote{\url{https://atcoder.jp/}} which ALE-Bench \citep{imajuku2025alebench} later used to evaluate AI systems.

To our knowledge, no earlier competition combines four features: one workflow must cover many problems at once, so that generality is rewarded over tuning for a single problem; on the Main track, the scored submission is algorithm code that the organizers run on hidden instances under fixed compute, rather than a formulation or a set of solutions; every solution passes an independent checker that recomputes the objective; and the final stage tests transfer to problems outside the published benchmark. The Self-hosted track scores uploaded solutions instead.

%=====================================================================
\section{Competition Format}
%=====================================================================

\paragraph{Teams} Participants enter through the competition portal as teams of one or more people. Each person holds only one account and belong to only one team at a time. A team chooses the Main or Self-hosted track when it is created. Registration and team formation close at 23:59 EST (UTC-5) on 27 December 2026. Teams may change their track before this deadline; afterward, their track is fixed for the competition.

\paragraph{Stage 1: Qualification} Stage 1 runs from 00:00 EST (UTC-5) on 4 January 2027 until 23:59 EST on 31 January 2027. Sixty problems are released in four weekly rounds of 15. The rounds open at 00:00 EST on 4, 11, 18 and 25 January, respectively, and each remains open for exactly seven days. A submission covers only the currently open round. Main-track submissions must be accepted by the portal before the round closes; their evaluation may finish afterward. Self-hosted uploads must meet the applicable instance, submission-window, and round deadlines. Teams follow a public leaderboard that accumulates results across rounds.

\paragraph{Main Track: Algorithm Design} Teams submit algorithm code, the scored submission, which we run in a sandbox with no network, 2~vCPU and 4~GB of memory. Only open-source dependencies are allowed. The overall champion comes from this track.

\paragraph{Self-hosted Track: Tool-Augmented Solving} Teams solve the instances on their own machines, by any means, including commercial solvers, and upload solutions within a timed window. The uploaded solutions are scored; we verify the solutions, not the process.

\paragraph{Qualification for Stage 2} Finalists are selected separately within each track using the final Stage 1 ranking across all 60 problems, after eligible submissions have been evaluated and eligibility checks have been completed. Teams are ranked by score S, highest first, and then by normalized time U, lowest first (Section~\ref{sec:metrics}). The number of finalist places in each track is TBA and will be announced when Stage 1 ends. If teams remain tied on both S and U at the qualification cutoff, all such tied teams advance, even if this increases the number of finalists.

\paragraph{Stage 2: Final} 

Stage 2 takes place on 8–14 February 2027 and introduces new problems outside the published benchmark, with unseen instances. Main-track finalists submit both their algorithm code and the source code of the agentic framework used to produce it. Self-hosted finalists submit solution files only. Final rankings are determined on a private final dataset. Detailed Stage 2 rules will be published on 1 February 2027, when Stage 1 ends, and will specify the exact development and submission deadlines, submission formats, and verification requirements.

\paragraph{Leaderboards} Each stage has one leaderboard per track; the two tracks are never compared. Each problem category and each formulation is also ranked on its own (Section~\ref{sec:data}). Baselines run by the organizers appear as unranked reference bands (Section~\ref{sec:baselines}).

%=====================================================================
\section{Datasets}
\label{sec:data}
%=====================================================================

\paragraph{Source} The problems come from the FrontierOR benchmark \citep{kong2026frontieror}: 180 problems derived from methodologically diverse papers in leading OR venues. Its audited release \citep{frontieror2026audited} covers all 180, with 1{,}080 instance/reference pairs, each reference accepted by the problem's checker. Every problem comes with a natural-language statement, instance and solution schemas, a feasibility checker, instances with $10^2$ to $10^7$ variables and constraints, and reference solutions computed with Gurobi. Each problem has a slug named after its source paper, such as \texttt{barnhart2000}.

\paragraph{Stage 1 Problems} The organizers select 60 of the 180 problems for Stage~1, none of them from the testing stage, chosen to cover diverse problem classes, formulations, objective directions and algorithm runtimes. Each problem has two labels (Table~\ref{tab:categories}): a \emph{category}, the application domain it comes from, and a \emph{formulation}, the mathematical form of the model behind it. A round's problems stay hidden until the round opens.

\begin{table}[htbp]
\floatconts{tab:categories}
  {\caption{Problem categories and formulations. Every problem has one of each. The examples are illustrative.}}
  {\small
   \begin{tabular}{@{}ll@{}}
   \toprule
   Category & Example problems \\
   \midrule
   Routing & vehicle routing variants, dial-a-ride, orienteering \\
   Production \& Inventory & lot sizing, production planning \\
   Scheduling & crew scheduling, machine scheduling \\
   Packing \& Partitioning & bin packing, cutting, set partitioning \\
   Graph \& Quadratic Optimization & graph partitioning, quadratic assignment \\
   Network Design \& Location & multicommodity flow, facility location \\
   \midrule
   Formulation & Model \\
   \midrule
   Combinatorial & discrete choices of routes, sequences, sets or assignments \\
   Mixed-integer & linear models with integer and continuous variables \\
   Quadratic & quadratic objective or constraints \\
   Stochastic & uncertain data given by scenarios or distributions \\
   Robust \& bilevel & worst case over an uncertainty set, or a leader and a follower \\
   \bottomrule
   \end{tabular}}
\end{table}

\paragraph{Instance Sets} Each problem has three disjoint sets of instances. \emph{Public} instances are published for development and accessible to registered teams; they are never scored. \emph{Private Main} instances score the Main-track submissions and are never published. \emph{Private Self-hosted} instances score the Self-hosted track; Self-hosted teams receive them during a submission, so they are kept apart from the Main set. Instances are generated iteratively: random instances are generated for each problem, their difficulty is evaluated with Gurobi, and generation continues until the set has the desired difficulty.

\paragraph{Reference Objectives} Each private instance has a reference objective $z^\ast$: a good checker-accepted objective. Reference solutions are computed with long-running Gurobi solves or specialized algorithms for the given problem. 

\paragraph{Example} Appendix~\ref{app:example} gives a complete problem from the testing stage, \texttt{barnhart2000}: its statement, instance schema and solution schema.

\paragraph{Contributions} Anyone registered may propose a new problem, or new instances for an existing problem. Each contribution should include a description, related work and files. Each contribution is reviewed individually and used to extend the benchmark. Accepted contributions are credited to the contributors.

%=====================================================================
\section{Evaluation Metrics}
\label{sec:metrics}
%=====================================================================

Each solution is checked, scored by its ratio to its instance's reference, and the instance scores are averaged into a submission score. Only private instances count.

\paragraph{Feasibility} The checker verifies every constraint and recomputes the objective from the decisions. A reported \texttt{objective\_value} that disagrees with the recomputed one beyond the objective tolerance is rejected; otherwise, the recomputed value $z$ is the one scored. In the testing stage and Stage~1, each problem's checker is published with it, as the evaluation runs it, so teams can check feasibility themselves; Stage~2 checkers are not published. Teams see the verdict for each instance.

\paragraph{Instance Score} A feasible solution with objective $z$ is scored against the reference objective $z^\ast$. A solution is \emph{better} than the reference when $z < z^\ast$ for minimization and $z > z^\ast$ for maximization. The score is
\begin{equation}
s = \begin{cases}
\min(z^\ast / z,\; 2) & \text{if minimizing and } z, z^\ast > 0, \text{ or maximizing and } z, z^\ast < 0,\\
\min(z / z^\ast,\; 2) & \text{if maximizing and } z, z^\ast > 0, \text{ or minimizing and } z, z^\ast < 0,\\
2 & \text{if } z z^\ast \le 0 \text{ and } z \text{ is better than } z^\ast,\\
1 & \text{if } z = z^\ast = 0,\\
0 & \text{if } z z^\ast \le 0 \text{ and } z \text{ is worse than } z^\ast.
\end{cases}
\label{eq:score}
\end{equation}
For positive objectives, minimization scores $z^\ast / z$ and maximization scores $z / z^\ast$; for negative objectives the two ratios swap. Either way, matching the reference scores 1, a worse solution less and a better one more, up to the cap of~2. When $z$ and $z^\ast$ have opposite signs, or either is 0, a ratio has no meaningful scale: a better solution scores 2, an equal one 1 and a worse one 0. An instance scores 0 when there is no feasible solution: the algorithm exceeded the time limit or the memory limit, crashed, wrote no readable solution, or the checker rejected it.

\paragraph{Tolerances} Each problem's checker sets its own tolerances. Constraints, including integrality, must hold within an absolute tolerance, typically $10^{-5}$. The reported \texttt{objective\_value} must agree with the recomputed objective within a relative tolerance, usually 0.5\%, depending on the problem. The score adds no tolerance of its own: $z = z^\ast$ only when the two values are identical, and only an objective of exactly 0 counts as zero.

\paragraph{Time Limits} The score rewards quality only, so speed is enforced through the time limits instead. Each problem falls into one of three time classes, with a limit of 10~seconds, 60~seconds or 5~minutes per instance. The classification is based on problem difficulty and application domain. The testing stage uses 60~seconds for every problem.

\paragraph{Submission Score} A submission covers the problems $P_r$ of one round~$r$. Let $I_p$ be the private instances of problem $p$ in the team's track; every one of them is scored. The submission's problem scores and its round score are
\begin{equation}
S_p = \frac{1}{|I_p|} \sum_{i \in I_p} s_i, \qquad R_r = \frac{1}{|P_r|} \sum_{p \in P_r} S_p .
\label{eq:round}
\end{equation}
Every problem weighs the same, however many instances it has. The denominator is always the whole round: a partial submission scores 0 on every instance of a problem it has no algorithm for (Self-hosted: on every instance without an uploaded solution).

\paragraph{Stage Score} In each round, exactly one submission counts: the team's completed submission with the highest $R_r$. All the round's problem scores come from that submission; a round's result is never assembled from the best problem scores of different submissions. Among submissions with equal $R_r$, the one with the lower time $U_r$ (below) counts, then the earliest. The stage score is the mean over the problems $P$ of all rounds opened so far,
\begin{equation}
S = \frac{1}{|P|} \sum_{p \in P} S_p = \sum_r \frac{|P_r|}{|P|} R_r ,
\label{eq:stage}
\end{equation}
each $S_p$ taken from the counted submission of its round. A round without a completed submission, including the open round before the team submits to it, has $R_r = 0$ and its problems stay in~$P$. Since $S$ grows with each $R_r$, the counted submissions are also those that maximize~$S$.

\paragraph{Ranking} Teams are ranked by $S$, highest first. Scores are compared without rounding, and rounded to four decimal places for leaderboard display. Equal scores are ranked by the time $U$, lowest first, and teams equal on both share a rank. $U$ is averaged like $S$, over the same instances and problems, from the time each instance used as a share of the most it could use:
\begin{gather}
u_i = \begin{cases}
\min(t_i / T_p,\; 1) & \text{if the solution to } i \text{ is feasible},\\
1 & \text{otherwise},
\end{cases}
\label{eq:time}\\
U_p = \frac{1}{|I_p|} \sum_{i \in I_p} u_i, \qquad U = \frac{1}{|P|} \sum_{p \in P} U_p , \nonumber
\end{gather}
and $U_r$ is the mean of $U_p$ over $P_r$. Here $t_i$ is the algorithm's wall time on the Main track and the time from download to upload on the Self-hosted track, and $T_p$ is the problem's time limit, plus the 60-second upload grace on the Self-hosted track. An instance without a feasible solution counts the full time whatever the reason: a time-out, a crash, an infeasible or rejected solution, a skipped instance, a problem without an algorithm or a missing upload. Every problem of a round without a submission has $U_p = 1$. Leaving work undone therefore never lowers $U$, and a fast failure ties with a slow one. A category or a formulation is ranked the same way, by the mean of $S_p$ and then of $U_p$ over its released problems.

\paragraph{Discussion} The ratio is simple to explain and normalizes each instance by its own reference, so problems with objectives of very different magnitude can be averaged. It has known limits, which we accept for the first edition: it is not invariant to a constant offset in the objective, so a problem whose objective is mostly a fixed cost compresses all teams near~1 and weighs less in the ranking than its equal weight suggests; and weak references inflate scores up to the cap.

%=====================================================================
\section{Evaluation Protocol}
%=====================================================================

\paragraph{Algorithm Contract} A Main-track submission is a zip or tar.gz archive of up to 4~MB with one folder per problem, named by its slug, each holding a \texttt{solve.py}. Each folder is mounted on its own, so algorithms cannot share code. The sandbox runs
\begin{center}
\begin{minipage}{0.92\linewidth}
\small
\begin{verbatim}
python solve.py --problem <slug> --instance /work/in/instance.json \
                --output /work/out/solution.json --time-limit <seconds>
\end{verbatim}
\end{minipage}
\end{center}
and expects a JSON object with a numeric \texttt{objective\_value} and the decisions the problem's solution schema defines. Algorithms are Python only; teams who want another language use the Self-hosted track.

\paragraph{Environment} Our images run Python 3.12 or 3.14 with NumPy, SciPy, pandas, NetworkX, OR-Tools, PuLP, HiGHS and Pyomo, and their full lockfiles are published. A team may add up to 50 open-source packages from PyPI, within the documented installation and resource limits; they are installed once before the first run, outside the time limit, and reused by later submissions with the same list.

\paragraph{Main-Track Execution} Each instance runs in its own gVisor\footnote{\url{https://gvisor.dev}} sandbox with no network, 2~vCPU, 4~GB of memory, 1~GB of disk, a read-only algorithm folder and the problem's time limit (Figure~\ref{fig:pipeline}). Each problem's public instances run first; if the algorithm fails on one, the problem's private instances are skipped and score 0. The checker then verifies each solution. Teams see each instance's score or the reason it scored 0, but no algorithm logs, objective values or run times for private instances, since these could reveal the instances.

\paragraph{Self-hosted Submissions} Opening a submission starts a time window long enough to solve every private instance of the round one after another, each within its time limit plus the grace below, plus 15~minutes. The team starts each private instance separately. The instance submission process consists of: starting the evaluation window, downloading the instance data, solving the instance locally, and uploading the solution. Each instance has its time limit (the same as on the Main track) plus 60~seconds of grace for the download and upload. Uploads are final. The submission closes when the last solution is uploaded, after which the checker and scoring run on our server, with the same scoring as the Main track.

\begin{figure}[htbp]
\floatconts{fig:pipeline}
  {\caption{Evaluation pipeline. Teams work through the portal; the compute service queues one job per instance and
   runs each in its own gVisor sandbox, and the scores flow back to the portal and the leaderboard. A Self-hosted submission takes
   the same path once its window closes, with only the checker running on the uploaded solutions.}}
  {\begin{tikzpicture}[
     section/.style={draw=archblue, thick, rounded corners=5pt, fill=archblue!6},
     item/.style={draw=archblue, rounded corners=3pt, fill=white, align=center, minimum width=3.7cm,
                  minimum height=0.85cm, inner sep=2pt},
     stage/.style={draw=archblue, dashed, rounded corners=2pt, minimum width=1.3cm, minimum height=0.6cm,
                  font=\small},
     title/.style={font=\bfseries, text=archblue!70!black},
     sub/.style={font=\scriptsize\itshape, text=black!60},
     arr/.style={-{Stealth[length=2mm]}, archblue, thick},
     lbl/.style={font=\scriptsize, text=black!65}
   ]
     % Team, outside the system
     \draw[thick] (0.85,0.72) circle (0.22);
     \draw[thick] (0.4,0.05) to[out=90, in=180] (0.85,0.42) to[out=0, in=90] (1.3,0.05);
     \node[font=\bfseries] at (0.85,-0.25) {Team};
     \node[sub] at (0.85,-0.62) {participants};

     % Portal
     \draw[section] (4.1,-2.4) rectangle (8.3,2.4);
     \node[title] at (6.2,2.05) {Portal};
     \foreach \y/\name/\what in {1.2/Dashboard/{team, submissions}, 0.2/API/{tokens, uploads},
                                 -0.8/Leaderboard/{stage, track, category}, -1.8/Problems \& docs/{statements, schemas}}
       \node[item] at (6.2,\y) {\small \name\\[-2pt]\scriptsize\itshape\color{black!60}\what};

     % Compute service
     \draw[section] (10.7,-2.4) rectangle (14.9,2.4);
     \node[title] at (12.8,2.05) {Compute service};
     \node[item] (queue) at (12.8,1.2) {\small Queue\\[-2pt]\scriptsize\itshape\color{black!60}one job per instance};
     \draw[draw=archblue, rounded corners=3pt, fill=white] (10.95,-2.2) rectangle (14.65,0.2);
     \node[font=\small] at (12.8,-0.08) {gVisor sandbox};
     \node[sub] at (12.8,-0.45) {no network, 2 vCPU, 4 GB};
     \node[stage] at (12.0,-1.45) {algorithm};
     \node[stage] at (13.6,-1.45) {checker};
     \draw[arr] (queue.south) -- (12.8,0.22);

     % Links between the sections
     \draw[arr] (1.75,0.25) -- node[lbl, above] {submissions} (4.05,0.25);
     \draw[arr] (4.05,-0.25) -- node[lbl, below] {scores, feedback} (1.75,-0.25);
     \draw[arr] (8.35,0.25) -- node[lbl, above] {jobs} (10.65,0.25);
     \draw[arr] (10.65,-0.25) -- node[lbl, below] {instance scores} (8.35,-0.25);
   \end{tikzpicture}}
\end{figure}
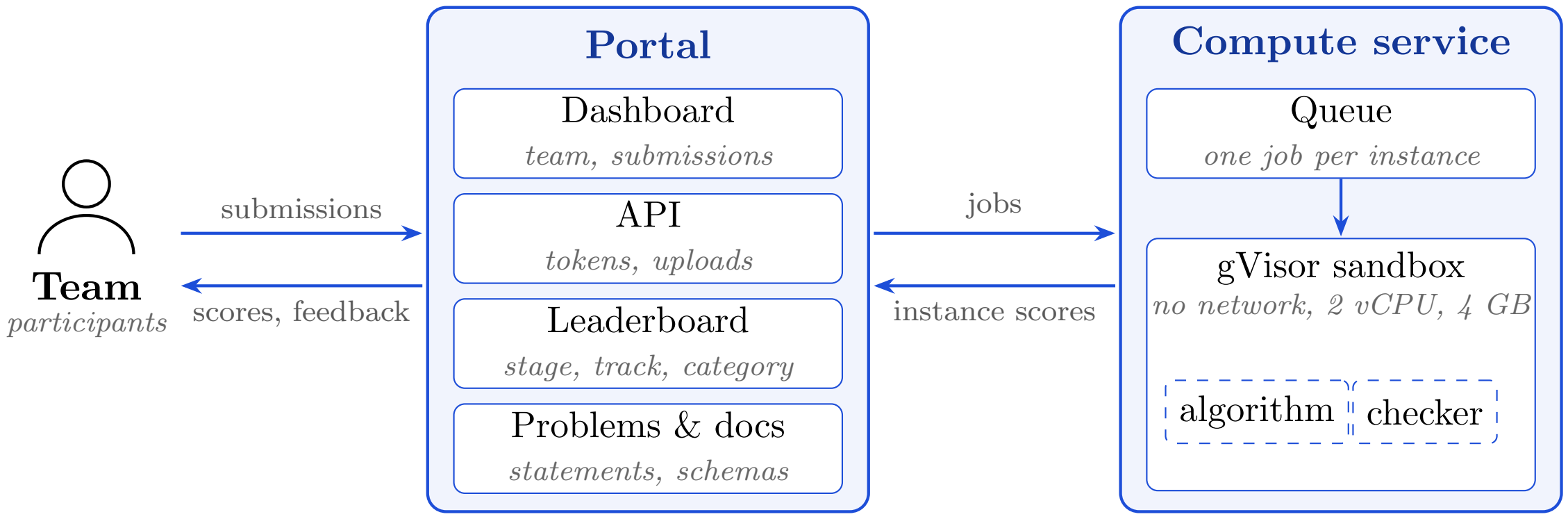

\paragraph{Infrastructure} The portal and API run as a web application backed by PostgreSQL and object storage. Submissions are queued and evaluated by an autoscaling fleet of CPU workers on AWS, each job receiving a whole worker's 2~vCPU and 4~GB. Failures on our side are retried; an instance that still fails after three tries scores~0. Each team has one submission per UTC day, four during the testing stage.

%=====================================================================
\section{Baselines and Starter Code}
\label{sec:baselines}
%=====================================================================

\paragraph{Starter Repository} A public starter repository\footnote{\url{https://github.com/Alpha-Z-Pte-Ltd/frontieror-challenge-starter}} works for both tracks. Its example agent receives a problem and must produce a function \texttt{solve(instance, time\_limit\_s)}: it asks a model for code, tests it on the public instances, and asks for fixes until it works. The prompt adapts FrontierOR's one-shot prompt to the challenge: open-source solvers instead of Gurobi, 2~CPUs, and a function instead of a whole program. A toolkit simplifies the process of downloading the problems, running the agent on every problem, testing the algorithms on the public instances (optionally in the sandbox's Docker image), and submitting the results. Competitors should further extend this starter code to build their own agentic workflows.

\paragraph{Baselines} The organizers run the starter code with several frontier models and submit the results through the same API and sandbox as any team. This results in the baseline solutions shown on the leaderboard. Baseline scores appear on the Stage~1 leaderboard as each round opens; a later version of this paper reports them.

%=====================================================================
\section{Rules}
%=====================================================================

\paragraph{Eligibility} Participants must hold only one account and belong to one team.

\paragraph{Dependencies} On the Main track, commercial solvers and remote services are not available in the
sandbox, and extra packages must be open source. Packages are not all checked automatically; prize-eligible
submissions may be audited. The Self-hosted track permits any software.

\paragraph{Fair Play} Algorithms code must not try to escape the sandbox, access the network, tamper with the
evaluation or read instances it was not given. Private instances are confidential: they must not be extracted,
reconstructed or shared, and Self-hosted teams must not pass their instances to anyone. Collusion between teams and
the use of several accounts or teams to gain submissions are prohibited. The organizers decide questions of scoring,
eligibility and interpretation, and may disqualify a submission or team that breaks the rules.

\paragraph{Intellectual Property} Teams keep ownership of their code and solutions. They grant the organizers a
non-exclusive, royalty-free license to store, execute, evaluate and analyze them in order to run and verify the
competition, and to publish scores and rankings. Submitted code and solutions are not published without the team's
permission; winning approaches may be described in aggregate or with consent.

\paragraph{Ethics and Accessibility} The problems come from published literature and contain no personal data.
Participation is free. On the Main track the organizers provide the evaluation compute, so a team needs only a laptop
and access to an LLM for its agent. The cost of LLM calls falls on the teams, which can favor well-funded groups.

%=====================================================================
\section{Timeline and Awards}
\label{sec:promotion}
%=====================================================================

\paragraph{Timeline} Table~\ref{tab:schedule} gives the schedule. The initial registration and testing stage is intended to let teams familiarize themselves with the portal and the submission process. This stage is also intended to let the organizers test the infrastructure and the scoring. Stage~1 releases the 60 competition problems used for qualification scoring in four weekly rounds of 15 problems.

\begin{table}[htbp]
\floatconts{tab:schedule}
  {\caption{Schedule. Stage dates are in US Eastern time.}}
  {\small
\begin{tabular}{
  @{}>{\raggedright\arraybackslash}p{0.34\linewidth}
     >{\raggedright\arraybackslash}p{\dimexpr0.66\linewidth-2\tabcolsep\relax}@{}
}\toprule
   Date & Milestone \\
   \midrule
   16 Sep 2026 & Registration and a public testing stage with six problems open \\
   27 Dec 2026, 23:59 EST & Registration, team formation, and testing close \\
   4 Jan 2027, 00:00 EST \newline to 31 Jan 2027, 23:59 EST & Stage~1 (qualification): four weekly rounds of 15 problems \\
   1 Feb 2027, when Stage 1 ends & Stage~2 rules published \\
   1 to 7 Feb 2027 & Break and Stage~1 wrap-up \\
   8 to 14 Feb 2027 & Stage~2 (final): development window and final evaluation \\
   TBA & Results, awards, artifacts; dated in the Stage~2 rules \\
   \bottomrule
   \end{tabular}}
   \textit{Note.} All timestamps shown above use EST (UTC$-5$). Submissions must be received before the applicable closing timestamp.
\end{table}

\paragraph{Awards} Table~\ref{tab:awards} lists the awards. The prizes go to the overall ranking, which comes from the Main track, for a pool of US\$21{,}000. Recognitions reward results and contributions beyond the ranking: the winner of each sub-track, that is each problem category and each formulation (Section~\ref{sec:data}), the best new problem or instance suite, and the best write-up of an agent's methodology, architecture and findings. Sub-track winners are announced with the results.

\begin{table}[htbp]
\floatconts{tab:awards}
  {\caption{Awards.}}
  {\small
   \begin{tabular}{@{}llr@{}}
   \toprule
   Award & Place & Prize (US\$) \\
   \midrule
   Winner & 1st overall & 10{,}000 \\
   Runner-up & 2nd overall & 5{,}000 \\
   Merit Awards & 3rd to 6th overall & 1{,}500 each \\
   Sub-track Winners & 1st per category and formulation & recognition \\
   Best New Problem / Instance & contribution & recognition \\
   Best Write-Up & write-up & recognition \\
   \bottomrule
   \end{tabular}}
\end{table}

%=====================================================================
\section{Organisation}
%=====================================================================

The challenge is organized by FrontierOR, MIT and Alpha Z. The FrontierOR team \citep{kong2026frontieror} provides the benchmark, its audit and the checkers; Alpha Z builds and runs the portal and the evaluation infrastructure. The executive committee is Hai Wang (chair), Chonghe Jiang, Minwei Kong, Ao Qu, Bretislav Hajek and Huatian Gong. The advisory committee is Cynthia Barnhart and Jinhua Zhao from MIT. The organizers provide the portal, the problems, instances and checkers, the starter repository and baselines, and support by email. Alpha Z sponsors the prize pool and the evaluation compute.

%=====================================================================
\section{Conclusion}
%=====================================================================

The goal of this competition is to find out whether agentic workflows can design OR algorithms that hold up on unseen data, across many problems and under a fixed budget. To that end, we propose 60 of the 180 audited FrontierOR problems in a qualification stage of weekly rounds, three time classes, new problems in a final stage, independent checkers, hidden instances and two tracks that separate algorithm design under fixed compute from tool-augmented solving. 

\acks{We thank the authors of the FrontierOR benchmark for the data and checkers, and Alpha Z for sponsoring the prize pool and the evaluation compute.}

\bibliography{white_paper}

\clearpage
\appendix

\section{Example Problem}
\label{app:example}

The testing stage's \texttt{barnhart2000}, Origin-Destination Integer Multicommodity Flow, comes from
\citet{barnhart2000}. Below are its statement and schemas as published on the problem page. The schemas are
annotated examples: each value gives its type in angle brackets, then its meaning, and a list shows one element.
The objective is minimized.

\paragraph{Statement}

A telecommunications network operator manages a directed network consisting of a set of nodes and a set of directed arcs connecting pairs of nodes. Each arc has a known capacity, expressed in units of bandwidth. The operator must handle a set of commodities, where each commodity represents a group of calls or messages sharing a common origin node and a common destination node. Each commodity has a known demand quantity representing the bandwidth it requires.

Every commodity must be routed as a single, unsplittable unit along exactly one directed path from its origin node to its destination node, or it may be rejected entirely. To allow rejection, the network includes, for each commodity, an artificial direct arc from the commodity's origin to its destination. The artificial arc has no capacity limit and carries a cost equal to the revenue that would have been earned by successfully routing that commodity. All physical arcs in the network have zero routing cost for every commodity.

The operator must choose, for each commodity, exactly one origin-to-destination path (which may be the artificial rejection arc or any path through the physical network). The total demand flowing through any physical arc, summed across all commodities whose chosen paths include that arc, must not exceed the arc's capacity. Flow must be conserved at every node for every commodity: at the origin node of a commodity, exactly one unit of that commodity's flow is produced; at the destination node, exactly one unit is absorbed; and at every intermediate node, the total flow of that commodity entering the node equals the total flow leaving it.

The goal is to minimize the total cost. For each combination of commodity and arc, the cost contribution equals the commodity's demand multiplied by the per-unit cost for that commodity on that arc, counted only when that arc lies on the commodity's chosen path. The total cost is the sum of these contributions across all commodities and all arcs. Because physical arcs have zero cost, the total cost reduces to the sum of artificial-arc costs for rejected commodities, which equals the total lost revenue from commodities not routed through the physical network. Equivalently, minimizing total cost maximizes the total revenue earned from commodities successfully carried on the physical network.

\paragraph{Instance Schema}

\begin{Verbatim}
{
  "network": {
    "num_nodes": "<int> Total number of nodes in the directed network.",
    "num_arcs": "<int> Total number of directed physical arcs in the network.",
    "nodes": "<list[int], length num_nodes> Identifiers for all nodes in the network.",
    "arcs": [
      {
        "arc_id": "<int> Unique identifier for this arc.",
        "from_node": "<int> Node where this directed arc originates.",
        "to_node": "<int> Node where this directed arc terminates.",
        "capacity": "<int> Maximum total bandwidth that can flow through this arc across all commodities.",
        "cost": "<int> Per-unit routing cost for any commodity using this arc."
      }
    ]
  },
  "commodities": {
    "num_commodities": "<int> Total number of commodities (call groups) to be routed or rejected.",
    "commodity_list": [
      {
        "commodity_id": "<int> Unique identifier for this commodity.",
        "origin": "<int> Node where this commodity originates.",
        "destination": "<int> Node where this commodity must be delivered.",
        "demand": "<int> Bandwidth required by this commodity.",
        "revenue": "<int> Revenue earned if this commodity is successfully routed through the physical network.",
        "artificial_arc_cost": "<int> Cost incurred if this commodity is rejected instead of being routed."
      }
    ]
  },
  "objective": "<str> Direction of optimization for the total routing and rejection cost."
}
\end{Verbatim}

\paragraph{Solution Schema}

\begin{Verbatim}
{
  "objective_value": "<float> Total cost of rejected commodities across the entire routing plan.",
  "commodities": [
    {
      "commodity_id": "<int> Identifier of the commodity whose routing outcome is described in this entry.",
      "rejected": "<bool> Whether this commodity is rejected and sent on its artificial arc instead of being routed through the network.",
      "path_arcs": [
        {
          "from": "<int> Origin node of an arc used on this commodity's routing path.",
          "to": "<int> Destination node of an arc used on this commodity's routing path.",
          "arc_id": "<int> Identifier of an arc used on this commodity's routing path."
        }
      ]
    }
  ]
}
\end{Verbatim}

\noindent On the public \texttt{tiny\_instance}, an example algorithm finds a feasible solution with objective 111 against a
reference of 64, which would score $64/111 \approx 0.5766$.

\clearpage
\section{Limits}
\label{app:limits}

Limits as of 20 November 2026.

\begin{table}[h!]
\floatconts{tab:limits}
  {\caption{Limits.}}
  {\small
   \begin{tabular}{@{}lr@{}}
   \toprule
   Limit & Value \\
   \midrule
   Submissions per team per UTC day & 1 (4 in the testing stage) \\
   Archive size (Main) & 4 MB \\
   Unpacked archive size (Main) & 256 MB \\
   Extra PyPI packages (Main) & 50 \\
   Environment build time (Main) & 10 min \\
   CPU and memory per instance (Main) & 2 vCPU, 4 GB \\
   Disk per instance (Main) & 1 GB \\
   Processes per instance (Main) & 256 \\
   Solution file (Main) & 16 MB \\
   Time limit per instance & 10 s, 60 s or 5 min per problem (60 s in testing) \\
   Stage~1 rounds & 4 rounds of 15 problems, 7 days each \\
   Submission window (Self-hosted) & $\sum$ (time limit + 60 s) + 15 min, rounded up to 5 min \\
   Upload grace (Self-hosted) & 60 s \\
   Network access (Main) & none \\
   \bottomrule
   \end{tabular}}
\end{table}

\end{document}